\documentclass{article}

\usepackage{amsmath,amssymb,mathtools}
\usepackage{bm}
\usepackage{graphicx}
\usepackage{booktabs}
\usepackage{microtype}
\usepackage{xcolor}
\usepackage[draft]{hyperref}

\usepackage{caption}
\usepackage{siunitx}
\usepackage{multirow}
\usepackage{mathrsfs}
\usepackage[nolist]{acronym}

\usepackage{algorithm}        
\usepackage{algorithmicx}     
\usepackage{algpseudocode}    
\usepackage{amsmath}
\usepackage{booktabs}

\usepackage{spconf}
\usepackage{enumitem}
\setlist{nosep, leftmargin=14pt}

\usepackage{mwe} 

\title{WaveFuse-AL: Cyclical and Performance-Adaptive Multi-Strategy Active Learning for Medical Images}

\name{Nishchala Thakur $^{ \dagger}$ 
\qquad Swati Kochhar$^{\dagger}$ \qquad Deepti R. Bathula $^{\dagger}$  \qquad  Sukrit Gupta$^{\star}$}

\address{$^{\dagger}$  Department of Computer Science \& Engineering, Indian Institute of Technology Ropar, India\\
$^{\star}$ Department of Biomedical Engineering, Indian Institute of Technology Ropar, India }

\begin{document}
\ninept

\begin{acronym}[nolist]
    \acro{IoU}{intersection over Union}
    
    \acro{CNN}{Convolutional Neural Network}
    \acro{ViT}{Vision Transformer}
    \acro{GNN}{Graph Neural Network}
    \acro{DL}{Deep Learning}
    \acro{VGG}{Visual Geometry Group-16}
    \acro{R50}{ResNet-50}
    \acro{D121}{DenseNet-121}
    \acro{EffNet}{EfficientNet-B0}
    \acro{AMLC}{Autoimmune Machine Learning Challenge}

\end{acronym}

\maketitle
\begin{abstract}
Active learning reduces annotation costs in medical imaging by strategically selecting the most informative samples for labeling. However, individual acquisition strategies often exhibit inconsistent behavior across different stages of the active learning cycle. We propose Cyclical and Performance-Adaptive Multi-Strategy Active Learning (WaveFuse-AL), a novel framework that adaptively fuses multiple established acquisition strategies—BALD, BADGE, Entropy, and CoreSet throughout the learning process. WaveFuse-AL integrates cyclical (sinusoidal) temporal priors with performance-driven adaptation to dynamically adjust strategy importance over time. We evaluate WaveFuse-AL on three medical imaging benchmarks: APTOS-2019 (multi-class classification), RSNA Pneumonia Detection (binary classification), and ISIC-2018 (skin lesion segmentation). Experimental results demonstrate that WaveFuse-AL consistently outperforms both single-strategy and alternating-strategy baselines, achieving statistically significant performance improvements (on ten out of twelve metric measurements) while maximizing the utility of limited annotation budgets.

\end{abstract}
\begin{keywords}
Active learning, Query strategy selection, Sinusoidal selection, Performance adaptation.
\end{keywords}
\section{Introduction}

Active learning reduces annotation costs by iteratively selecting the most informative samples for labeling \cite{b1} by repeatedly picking  unlabeled images using defined query strategies such as uncertainty sampling, representativeness, or diversity and add them to the training set. This approach helps deep models achieve high accuracy with fewer labeled images, directly addressing the cost and expertise bottleneck in clinical datasets. Despite widespread AL adoption, many biomedical applications rely on a single query strategy (e.g., uncertainty, entropy, or representative selection) for all rounds of sample acquisition. Reusing one strategy across the annotation process may repeatedly select visually or statistically similar samples, causing over collection from particular regions of the feature space. For example, entropy based sampling can keep focusing on ambiguous but statistically similar images, failing to diversify the representation of the labeled set especially in imbalanced or heterogeneous clinical datasets resulting in poor generalization and downstream performance. 

Jung et al.~\cite{b7} proposed alternating acquisition functions to balance exploration and exploitation, demonstrating that cycling between diverse strategies can outperform fixed approaches by adapting to an evolving label set. However, stepwise alternation schemes suffer from abrupt regime shifts, leading to `thrashing' in sample scores and instability when strategies disagree on candidate importance. Motivated by these insights, we propose a novel  Cyclical and Performance adaptive Multi strategy Active Learning  that smoothly weights multiple acquisition functions across rounds using phase-shifted sinusoidal curves. 

Our approach enables a continuous and predictable mixing of uncertainty-based, diversity-based and hybrid strategies. We blend these strategies to avoid abrupt switches while adapting the exploration-exploitation balance throughout the learning cycle.
Our key contributions are:
\begin{enumerate}
    \item A novel adaptive multi-strategy framework that assigns phase-shifted sinusoidal priors to multiple acquisition strategies (BALD, BADGE, Entropy, CoreSet), enabling smooth temporal cycling of strategy importance across the active learning process.
    \item A unified weighting mechanism that fuses temporal priors with performance-based adaptation, enabling a smooth shift from early exploration to later performance-driven selection while preserving sample diversity.
    \item Comprehensive experiments across multiple medical imaging benchmarks—covering binary and multi-class classification, segmentation, varying annotation budgets (10-20\(\%\)), and explainable AI analyses to demonstrate the robustness and effectiveness of the proposed method.
\end{enumerate}

 \begin{figure*}[!t]
  \centering
  \includegraphics[width=\textwidth]{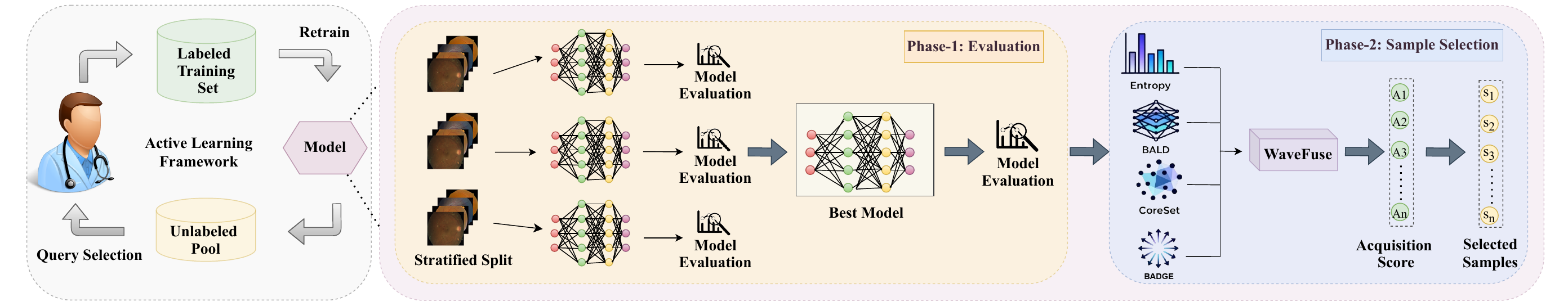 }
  {\caption{Overview of the proposed active learning framework for medical image analysis. In \textbf{Phase-1}, stratified splits are used to evaluate candidate models, selecting the best performing model. In \textbf{Phase-2}, sample selection is performed using a fusion of query strategies (Entropy, BALD, CoreSet, BADGE) combined via WaveFuse, producing acquisition scores for the selection of informative samples for labeling.}
}
  \label{fig:pipeline}
\end{figure*}
\section{Methodology}
\subsection{Problem Formulation}
Let \(\mathcal{D} = \{(\boldsymbol{X}_i, y_i)\}_{i=1}^{N}\) denote a dataset where \(\boldsymbol{X}_i\) is an image and \(y_i\) its label (class for classification, mask for segmentation). At round \(t\), we maintain a labeled set \(\mathcal{L}_t\) and an unlabeled pool \(\mathcal{U}_t\) where \(\mathcal{L}_t \cup \mathcal{U}_t = \mathcal{D}_{\text{train}}\) and \(\mathcal{L}_t \cap \mathcal{U}_t = \emptyset\). A holdout test set \(\mathcal{D}_{\text{test}}\) remains isolated.We initialize \(\mathcal{L}_0\) with \(|\mathcal{L}_0|\) stratified samples. At each round, the model \(f\) is trained on \(\mathcal{L}_t\), and a batch \(\mathcal{B}_t\) of \(b\) informative samples is selected from \(\mathcal{U}_t\). The pools are then updated, and the process repeats for \(T \) rounds. Multiple strategies have been proposed for selecting samples that can be be labeled from \(\mathcal{U}_t\).

\subsection{Established Selection Strategies}
\textbf{BADGE} \cite{b4}: Selects samples that are both informative and diverse by leveraging gradient-based representations. For a classifier \(f\), the gradient embedding of an unlabeled sample \(\boldsymbol{X}\) is given by
\begin{small}
\begin{equation}
   g_{\boldsymbol{X}} = (\hat{y} - f(\boldsymbol{X})) \otimes h(\boldsymbol{X}),   
\end{equation}
\end{small}
\noindent where \(\hat{y}\in\{0,1\}^{K}\) denotes the pseudo-label, \(h(\boldsymbol{X})\in\mathbb{R}^{D}\) is the penultimate-layer's input embedding, and \(\otimes\) denotes the outer product. For segmentation, BADGE uses global average pooled features \(z_{\boldsymbol{X}} = G(h(\boldsymbol{X})) \), where \(G(.)\) denotes global average pooling over spatial dimensions. Finally, it performs $k$-MEANS++ clustering over the set of gradient (\(g_{\boldsymbol{X}}\)) or pooled (\(z_{\boldsymbol{X}}\)) embeddings to select a batch of representative samples from the unlabeled pool, balancing uncertainty and diversity.\\


\noindent\textbf{Entropy}\cite{b2}: For an unlabeled sample $\boldsymbol{X}$, let $f_c(\boldsymbol{X})$ be the predicted class probability for class $c$. The \emph{entropy} is given by:
\begin{small}
\begin{equation}
    H(\boldsymbol{X}) = -\sum_c f_c(\boldsymbol{X}) \log f_c(\boldsymbol{X})
    \end{equation}
\end{small}
\noindent This strategy favors samples with high predictive uncertainty, encouraging the model to learn from ambiguous instances.\\


\noindent \textbf{BALD} \cite{b5}: Quantifies epistemic uncertainty via MC dropout with \(M\) stochastic forward passes. The mutual information, \(I(x)\), between predictions and model parameters (the BALD score) is given by:
\begin{small}
\begin{equation}
I(x) = H\left[\frac{1}{M}\sum_{m=1}^M f_m(y)\right] - \frac{1}{M}\sum_{m=1}^M H[f_m(y)]
\end{equation}
\end{small}
where
\(f_m\) is the predictive categorical distribution (softmax) under the \(m\)-th dropout realization and \(M\) is the number of MC dropout passes. For segmentation, we compute BALD per pixel, \(u\), and average spatially:
\begin{small}
\[{BALD}(x) = \frac{1}{U} \sum_{u=1}^{U} I_u(x),\quad U=H_{\text{img}} \times W_{\text{img}}.\]
\end{small}\\
\noindent \textbf{CoreSet} \cite{b3}: Formulates batch selection as a geometric covering problem to ensure selected samples are both diverse and representative of the entire data set. The method treats batch selection as the $k$-Center problem:
Given $h(\boldsymbol{X})$ for all unlabeled points $\boldsymbol{X} \in \mathcal{U}$, CoreSet selects a batch $\mathcal{B}$ of size $b$ minimizing the maximum distance from any point in $\mathcal{U}$ to its nearest selected sample:
\begin{equation}
\mathcal{B}^* = \underset{\mathcal{B} \subset \mathcal{U},\ |\mathcal{B}|=b}{\operatorname{arg\,min}}\, \max_{\boldsymbol{X} \in \mathcal{U}} \min_{\boldsymbol{X}' \in \mathcal{L} \cup \mathcal{B}} \| h(\boldsymbol{X}) - h(\boldsymbol{X}') \|_2
\end{equation}
\subsection{Proposed Sinusoidal Strategy Selector}
Our method introduces a novel adaptive mechanism that fuses sinusoidal temporal weighting with performance-driven feedback to optimize strategy selection.

\begin{figure}[ht]
  \includegraphics[width=0.9\columnwidth
  ]{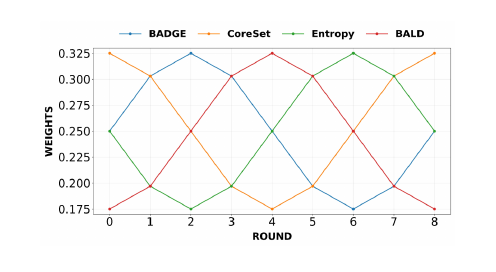 }
  \caption{Sinusoidal Strategy Weights across rounds}
  \label{fig:strategy_weights}
\end{figure} 

 The temporal prior assigns each strategy a phase-shifted sinusoid over $T$ rounds to yield gradual, predictable shifts in dominance - balanced exploration without abrupt alternation. For \(S\) strategies, each strategy \(s \in \{1, \ldots, S\}\) receives a phase-shifted sinusoid (Figure~\ref{fig:strategy_weights}):

\begin{figure*}[!h]
  \includegraphics[width=1\textwidth]{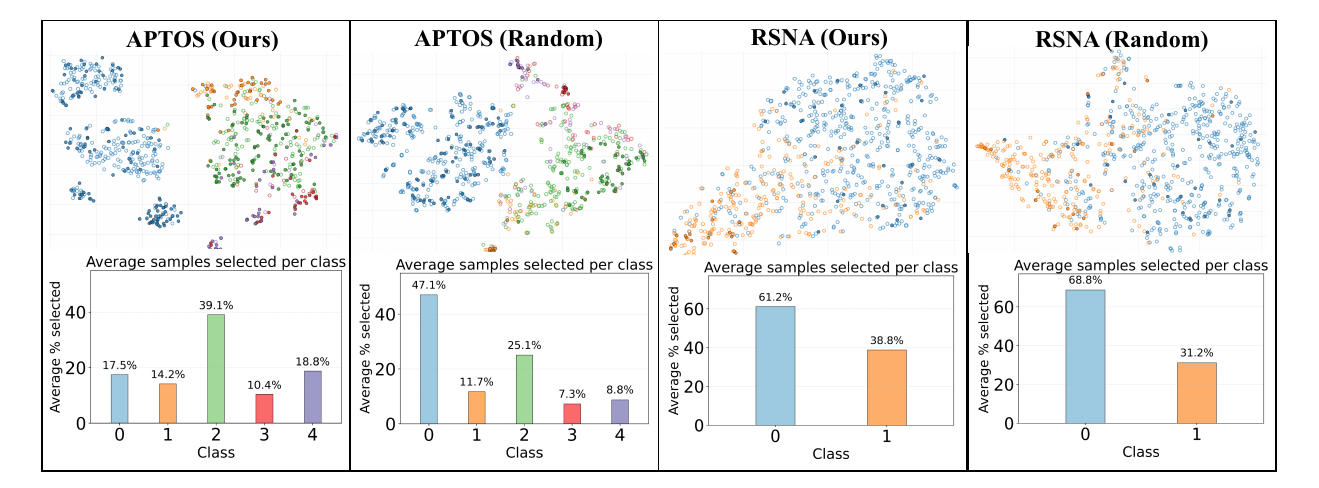 }
 \caption{t-SNE embeddings (top) and class-wise acquisition rates (bottom) comparing the proposed active-learning strategy with random sampling on APTOS-2019 (Class 0:\emph{No DR}-49.3\%, Class 1: \emph{Mild DR}-10.1\%, Class 2:\emph{Moderate DR}- 27.3\%, Class 3: \emph{Severe DR}- 5.3\%, and Class 4: \emph{Proliferative DR}- 8.1\%) and RSNA Pneumonia (Class 0: Normal - 67.73\% and Class 1: Pneumonia - 32.27\%). The labeled points are shown as colored solid circles, while unlabeled points are transparent circles with a solid border. Ours shows broader cluster coverage and reduced class bias.}
  \label{fig:tsne}
\end{figure*}
\begin{small}
   \begin{equation}
\Psi_s(t) = \sin\left(2\pi \frac{t}{T} + \phi_s\right) + 1, \quad \phi_s = \frac{2\pi s}{S}
\end{equation} 
\end{small}

\noindent where \(t\) is current round of active learning. In parallel, we maintain an exponentially smoothed, task-appropriate performance trace for each strategy \( s\), given by
\begin{small}
   \begin{equation}
\Omega_s(t) = \beta \cdot {\omega}_s(t) + (1-\beta) \cdot \Omega_s(t-1)
\end{equation} 
\end{small}

\noindent where  \(\omega_s(t)\) denotes the most recent performance metric (e.g., accuracy, F1, IoU, DICE, et cetera) observed after round $t$ and $\beta\in[0,1]$ is the smoothing factor controlling responsiveness versus stability (larger $\beta$ adapts faster but is more variance-sensitive, smaller $\beta$ is stabler but slower). The sinusoidal temporal prior, \(\Psi\), is combined with performance-based weighting, \(\Omega\), as 
\begin{small}
  \begin{equation}
    w_s(t) = \alpha \cdot \Psi_s(t) + (1-\alpha) \cdot \text{softmax}\left(\frac{\Omega_s(t)}{\tau}\right)
\end{equation}  
\end{small} 



\noindent where $\alpha\in[0,1]$ trades smoother prior cycling (larger $\alpha$) against performance reactivity (smaller $\alpha$), and $\tau>0$ controls the sharpness of the softmax (lower $\tau$ accentuates winners; higher $\tau$ spreads weights for stability). The final weights are normalized across all strategies such that their sum equals $1$. The batch budget $b$ is then apportioned proportionally via $q_s=\lfloor b\,w_s(t)\rfloor$. This strategy yields smooth, interpretable weight trajectories that remain adaptively focused on empirically strong strategies while retaining diversity and avoiding abrupt regime shifts.




\begin{table*}[!h]
\centering
\caption{Results of Strategies Across APTOS-2019, RSNA Pneumonia, and ISIC 2018 at 10\% and 20\% Label Budgets (Mean$_{Std}$). * denotes statistically significant improvements over the second-best baseline (paired t-test, $p < 0.05$)}
\label{tab:combined_results}
\scriptsize
\setlength{\tabcolsep}{3pt}
\begin{tabular}{@{}lcccccccccccc@{}} 
\toprule
\multirow{2}{*}{\textbf{Method}} &
\multicolumn{2}{c}{\textbf{APTOS 10\%}} &
\multicolumn{2}{c}{\textbf{APTOS 20\%}} &
\multicolumn{2}{c}{\textbf{RSNA 10\%}} &
\multicolumn{2}{c}{\textbf{RSNA 20\%}} &
\multicolumn{2}{c}{\textbf{ISIC 10\%}} &
\multicolumn{2}{c}{\textbf{ISIC 20\%}} \\
\cmidrule(lr){2-3} \cmidrule(lr){4-5} \cmidrule(lr){6-7} \cmidrule(lr){8-9} \cmidrule(lr){10-11} \cmidrule(lr){12-13}
& Acc. & F1 & Acc. & F1 & Acc. & F1 & Acc. & F1 & Dice & IoU& Dice& IoU\\
\midrule
Random\cite{b1} & $\underline{61.99}_{4.93}$ & ${60.49}_{4.76}$ & $\underline{66.05}_{4.41}$ & $\underline{66.73}_{4.33}$ & $72.63_{0.08}$ & $72.64_{0.09}$ & $\underline{80.28}_{0.41}$& $\underline{80.68}_{0.38}$& $71.92_{2.35}$ & $59.63_{2.22}$ & $73.92_{2.22}$& $61.63_{2.35}$\\
BADGE\cite{b4} & $60.38_{0.83}$ & $\underline{60.52}_{1.32}$ & $60.32_{1.84}$ & $60.10_{1.76}$ & $64.18_{1.57}$ & $63.27_{0.07}$ & $71.63_{1.20}$& $71.82_{1.14}$& $71.64_{3.26}$ & $58.94_{2.52}$ & $73.64_{2.52}$& $60.66_{3.26}$\\
BALD\cite{b5} & $60.99_{1.67}$ & $58.96_{3.74}$ & $64.38_{1.74}$ & $64.20_{2.13}$ & $71.22_{3.74}$ & $71.06_{3.71}$ & $76.31_{2.01}$& $76.25_{2.09}$& $73.27_{1.79}$ & $60.70_{1.75}$ & $75.27_{1.75}$& $62.70_{1.79}$\\
CoreSet\cite{b3} & $51.49_{1.49}$ & $50.95_{2.01}$ & $61.37_{3.09}$ & $60.64_{3.59}$ & $\underline{73.95}_{1.87}$ & $\underline{73.51}_{1.60}$ & $74.82_{1.34}$ & $74.52_{1.28}$& $73.59_{1.93}$ & $61.13_{1.87}$ & $75.59_{1.87}$& $63.13_{1.93}$\\
Entropy\cite{b2} & $39.48_{3.93}$ & $34.58_{4.32}$ & $34.74_{5.98}$ & $30.76_{0.93}$ & $71.22_{1.38}$ & $70.78_{1.68}$ & $69.60_{1.41}$& $69.60_{1.49}$& $71.78_{1.00}$ & $59.07_{1.11}$ & $73.78_{1.11}$& $61.07_{1.00}$\\
Margin\cite{b1} & $45.47_{2.32}$ & $42.10_{1.96}$ & $42.73_{2.84}$ & $37.70_{4.52}$ & $70.80_{2.62}$ & $70.23_{2.68}$ & $70.60_{2.33}$& $70.38_{2.41}$& $\underline{74.31}_{3.45}$ & $\underline{61.66}_{1.95}$ & $\underline{76.31}_{1.95}$& $\underline{63.66}_{3.45}$\\
Alternating\cite{b7} & $48.49_{4.25}$ & $42.21_{4.42}$ & $56.16_{4.6}$ & $59.93_{3.06}$ & $58.56_{2.86}$& $59.07_{1.98}$& $67.25_{2.29}$& $67.33_{2.27}$& $73.01_{2.47}$ & $58.63_{1.84}$ & $75.01_{1.27}$& $62.33_{3.02}$\\
\midrule
\textbf{Ours} & $\mathbf{65.24}_{0.47}$ & $\mathbf{65.04}_{0.90}^{*}$ & $\mathbf{77.45}_{1.46}^{*}$ & $\mathbf{76.03}_{1.53}^{*}$ & $\mathbf{76.50}_{1.59}^{*}$ & $\mathbf{76.40}_{1.66}^{*}$ & $\mathbf{83.27}_{1.44}^{*}$& $\mathbf{82.65}_{1.59}^{*}$& $\mathbf{77.57}_{1.38}$ & $\mathbf{65.97}_{1.19}^{*}$ & $\mathbf{79.57}_{1.19}^{*}$& $\mathbf{67.97}_{1.38}^{*}$\\
\bottomrule
\end{tabular}
\end{table*}

\section{Experimental Results}

\subsection{Datasets}
The APTOS-2019\cite{b8} and RSNA \cite{b9}datasets were used for classification tasks, while ISIC-2018 \cite{b10} was employed for segmentation. The APTOS-2019 dataset comprises 3,662 retinal fundus images categorized into five diabetic retinopathy (DR) severity levels:
\emph{No DR}-1,805 images; 49.3\%,
\emph{Mild DR}- 370 images; 10.1\%,
\emph{Moderate DR}- 999 images; 27.3\%,
\emph{Severe DR}- 193 images; 5.3\%, and
\emph{Proliferative DR}- 295 images; 8.1\%.
The RSNA Pneumonia Detection Challenge dataset includes 30,227 chest X-rays, with 20,482 normal (67.73\%) and 9,745 pneumonia (32.27\%) cases.

The ISIC-2018 dataset contains \textbf{2,594 dermoscopic images} with pixel-level lesion masks. 
RSNA images are of size $1024 \times 1024$, while APTOS and ISIC images vary in resolution.
We started with initial labeled sets consisted of \textbf{40 (APTOS)}, \textbf{300 (RSNA)}, and \textbf{40 (ISIC)} samples, with per-round query budgets of \textbf{40}, \textbf{300}, and \textbf{30}, respectively, for \textbf{10 active learning rounds}. This resulted in final labeled sets of \textbf{400 (10.92\%)}, \textbf{3,000 ($\sim$10\%)}, and \textbf{310 (11.9\%)} samples.
\begin{figure*}[ht]
  \centering
  \includegraphics[width=\textwidth, height = 3.7cm]{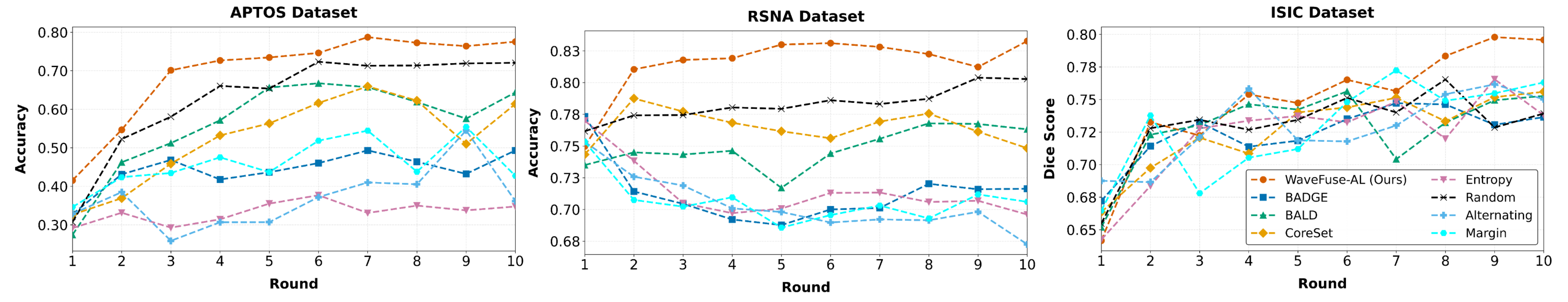}
\caption{Performance comparison of active-learning strategies across three datasets (across all the rounds) where we select \(~20\%\) of the total samples after 10 rounds. Our method consistently outperforms baselines across all datasets.}
  \label{fig:curves}
\end{figure*}
\subsection{Model and training configuration}
We use ImageNet-pretrained DenseNet-121 as backbone: for classification, a linear layer projects pooled encoder features to $C$ classes; for segmentation, DenseNet-121 serves as encoder in a U-Net-style decoder with skip connections. Training uses Adam optimizer ($\eta = 10^{-4}$), batch size $8$, $80$ epochs per model, and augmentation (random flips, rotations $\pm 10^{\circ}$, intensity adjustments). Classification loss is cross-entropy; segmentation uses Dice. Images are resized to $224 \times 224$ (classification) or $256 \times 256$ (segmentation) and normalized via ImageNet statistics. The sinusoidal strategy controller employs $\alpha = 0.3$ (decaying to $0.02$) for prior strength, weight bounds $[0.05,\,0.8]$ (floor/cap), $\beta = 0.30$, softmax temperature $\tau = 0.7$ (annealing to $\tau_{\text{low}} = 0.25$), dominance threshold $0.6$, and exploration $\epsilon = 0.10$ (decaying to $0.02$). The parameters are selected empirically. Experiments were performed on a NVIDIA RTX A2000 12GB GPU. All methods are evaluated using stratified cross-validation, results are reported as mean $\pm$ standard deviation across all runs.

\subsection{Model performance comparison}
Across datasets and label budgets, our method consistently outperforms strong active learning baselines on classification (APTOS, RSNA) and segmentation (ISIC) metrics, with larger gains at 20\% labeling where variance is lower. Results are reported in table \ref{tab:combined_results} as mean$_{Std}$, best scores are bolded, and statistical significance versus the best baseline is indicated by an asterisk using paired t-tests at \(p<0.05\). At 10\% labels, improvements are modest yet reliable, at 20\% labels, our approach achieves top Accuracy/F1 (APTOS, RSNA) and Dice/IoU (ISIC), reflecting superior sample efficiency under larger query budgets.

\subsection{Qualitative analysis of the results}
t-SNE embeddings  with class-wise acquisition rates in fig- \ref{fig:tsne} show that the proposed strategy achieves broader cluster coverage and lower class bias than random sampling on APTOS-2019 and RSNA. In APTOS, selections concentrate near inter-cluster margins and minority DR grades instead of over-sampling dense Class 0 cores. For RSNA, selections track the normal-pneumonia boundary instead of class cores, indicating boundary-focused sampling with improved coverage of the feature manifold. Due to space, ISIC visuals are omitted, but they exhibit the similar pattern.
\section{conclusion}

We presented WaveFuse-AL, a cyclical and performance-adaptive active learning framework that dynamically fuses multiple acquisition strategies through sinusoidal temporal priors and performance-based weighting. Experiments across diverse medical imaging tasks show that WaveFuse-AL delivers statistically significant improvements over single-strategy and static baselines while requiring only a small fraction of labeled data. These results highlight its robustness and effectiveness for reducing annotation costs in medical imaging.

\bibliographystyle{IEEEbib}
\bibliography{references}
\section{Compliance with ethical standards}
\label{sec:ethics}
This research study was conducted retrospectively using human subject data made available in open access. Ethical approval was not required as confirmed by the license attached with the open-access data.
\end{document}